\documentclass[runningheads]{llncs}

\usepackage{eccv}

\usepackage{eccvabbrv}

\usepackage{graphicx}
\usepackage{booktabs}
\usepackage{wrapfig}
\usepackage{graphicx}
\usepackage{caption}

\usepackage[accsupp]{axessibility}  

\usepackage{hyperref}

\usepackage{orcidlink}

\begin{document}

\title{AquaStereo: Enabling Underwater Stereo Matching via Depth-Conditioned Diffusion and Geometry Self-Distillation} 

\titlerunning{AquaStereo for Underwater Stereo Matching}

\author{
Qizhe Wei\textsuperscript{*}\inst{1}\orcidlink{0009-0006-9649-2610}
\and
Yingping Liang\textsuperscript{*}\inst{1}\orcidlink{0000-0001-5385-0015}
\and
Shaodi You\inst{2}\orcidlink{0000-0001-8973-645X}
\and
Ying Fu\textsuperscript{\textdagger}\inst{1}\orcidlink{0000-0002-6677-694X}
}

\authorrunning{Q. Wei et al.}

\institute{
Beijing Institute of Technology, Beijing, China\\
\and
University of Amsterdam, Amsterdam, The Netherlands \\
\email{\{qizhewei,liangyingping,fuying\}@bit.edu.cn}\qquad
\email{s.you@uva.nl}
}

\maketitle

\begingroup
\renewcommand{\thefootnote}{\fnsymbol{footnote}}
\footnotetext[1]{Equal contribution.}
\endgroup

\begingroup
\renewcommand{\thefootnote}{\textrm{\textdagger}}
\footnotetext[1]{Corresponding author.}
\endgroup

\begin{abstract}
Learning-based stereo matching models struggle in underwater environments due to scarce in-domain data and the difficulty of extracting discriminative correspondences from degraded imagery. In this work, we present \textbf{AquaStereo}, a perception-enhanced framework with a data simulation pipeline and a self-distillation strategy that jointly address data scarcity and feature degradation in underwater stereo matching. First, a depth-conditioned diffusion pipeline renders underwater stereo pairs while preserving binocular geometry, with a lightweight left-right consistency module ensuring geometric alignment. Training on this synthetic corpus effectively narrows the terrestrial–underwater gap and improves zero-shot robustness. Second, a frozen binocular teacher trained on clean terrestrial pairs guides a student exposed to rendered underwater pairs with perturbations. A stage-weighted sequence loss is performed to align the student's disparities with the teacher's geometry, while a clean-branch supervision with shared pseudo targets prevents scale drift. To further enhance feature stability under turbidity and low texture, we introduce learnable perception frames, a perception-enhanced feature formulation that constructs robust matching descriptors by fusing temporal cues from two auxiliary views encoded by a video backbone with semantic features extracted by a strong image encoder. Extensive experiments demonstrate that \textbf{AquaStereo} substantially improves robustness and zero-shot generalization in challenging underwater scenarios. The code is available at \url{https://github.com/qz-wei/AquaStereo}.
\end{abstract}

\section{Introduction}
\label{sec:intro}

\begin{figure*}[t]
  \centering
  \includegraphics[width=\textwidth,height=55mm]{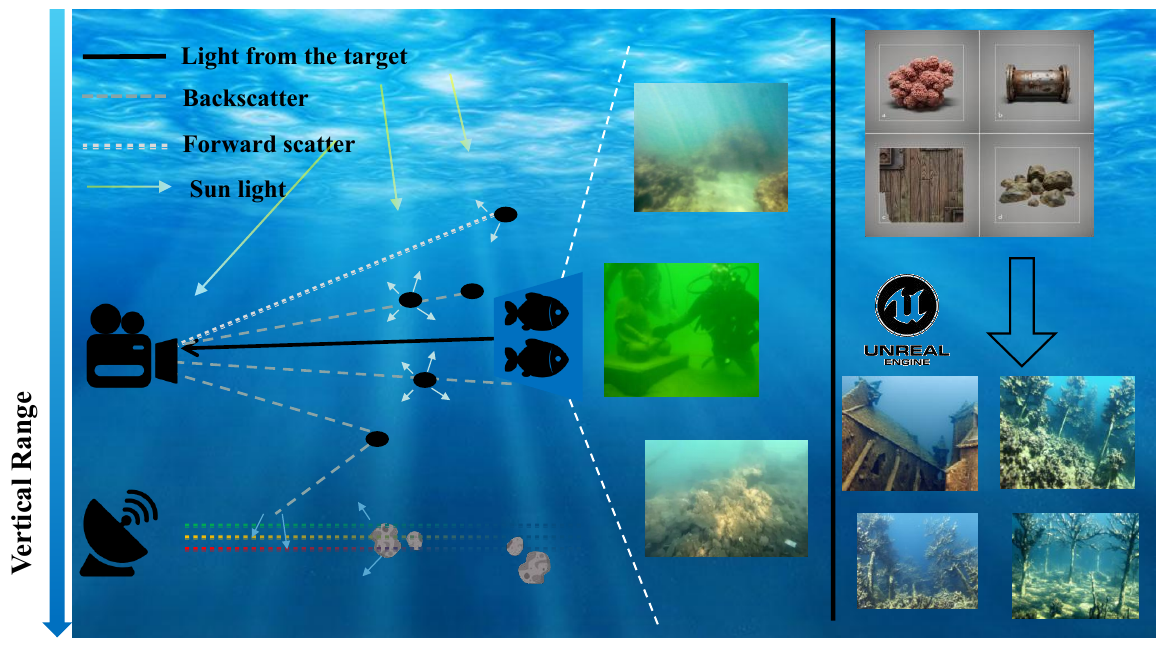}
\caption{\textbf{Principles of underwater data acquisition.}
\textbf{Left:} Underwater sensing with LiDAR and vision cameras. 
\textbf{Right:} CG-based underwater image rendering.}
  \label{fig:yuanli}
\end{figure*}
Stereo matching is a fundamental task in computer vision, spanning autonomous driving, robotics, 3D reconstruction, and virtual reality~\cite{scharstein2002taxonomy,hirschmuller2008stereo,zbontar2016stereo}. Accurate underwater depth is likewise critical for autonomous underwater vehicles and marine applications~\cite{paull2013auv,yuh2001underwater,gibson2016review,bailey2008archaeology,coleman2000underwater}. Despite remarkable progress in terrestrial stereo matching~\cite{chang2018psmnet,guo2024stereo,lipson2021raft,guo2023openstereo}, direct transfer to underwater scenes remains ineffective because underwater image formation differs fundamentally from terrestrial settings. Attenuation, scattering, and backscatter reduce contrast and distort color, violating common photometric assumptions and suppressing reliable textures. More critically for stereo, these degradations are often view-dependent, breaking left--right consistency and making cost-volume matching ambiguous. A key bottleneck is data. As illustrated in Figure~\ref{fig:yuanli}, underwater stereo data are typically obtained either by real-world capture or by CG rendering, yet both have clear limitations. Real capture is expensive and hard to scale; moreover, scattering and refraction degrade both passive vision and active ranging (e.g., LiDAR), often resulting in sparse or unreliable depth/disparity supervision (e.g., HIMB~\cite{Skinner2019UWstereoNet}, FLSea~\cite{Randall2023FLSea}, Squid~\cite{berman2020underwater}). In contrast, CG rendering/simulation provides dense supervision at low cost (e.g., VAROS~\cite{Zwilgmeyer2021VAROS}, UWStereo~\cite{UWStereo2024}), but lacks photorealism and introduces a noticeable domain gap to real underwater imagery.


To tackle the challenge of scarce and unreliable underwater stereo data, we present \textbf{AquaStereo}. We take a third route: generation. Compared with real capture and CG rendering, diffusion-based generation~\cite{Song2021DDIM,zhang2023controlnet,fang2023dual} can produce more realistic underwater appearances while remaining highly controllable and scalable. However, underwater stereo generation is more challenging than single-image translation~\cite{zhang2019ganet,Hambarde2021UWGANSD,li2023joint} or generic image generation~\cite{Rombach2022LDM,zhang2024deep}: it must preserve left--right correspondence and epipolar consistency under complex and highly variable underwater degradations. Violating this structure directly corrupts cost-volume matching in modern stereo pipelines~\cite{chang2018psmnet,xu2025igev++}. To control these non-deterministic appearance changes, we build a physics-inspired prompt pool encoding water types and degradation cues~\cite{Sea-Thru,chen2026survey}, constructed from CLIP-retrieved~\cite{clip} underwater imagery and LLM-curated underwater descriptions. We further incorporate a lightweight left--right consistency module to enforce binocular alignment, ensuring generated pairs remain suitable for downstream stereo training.

Building on the generated data, we further improve robustness via cross-domain self-distillation:
a frozen teacher provides stable geometric supervision, while the student learns from underwater pairs with perturbations;
clean-branch supervision with shared pseudo targets prevents scale drift
\cite{self-dis,zhang2022selfdistillation}.
Finally, we introduce a perception-enhanced matcher to stabilize features under turbidity and low-texture conditions by
concatenating two learnable auxiliary views with the stereo pair, encoding them with a video backbone, and fusing
image-level semantics from an image encoder to produce robust matching descriptors
\cite{change3d,oquab2023dinov2}.

Extensive experiments demonstrate that our pipeline significantly boosts underwater stereo depth estimation accuracy. In summary, our contributions are: (1) we propose \textbf{AquaStereo}, an underwater stereo matching framework that improves generalization and accuracy without target-domain fine-tuning; (2) we introduce a depth-conditioned diffusion pipeline with a physics-inspired prompt pool and a lightweight left--right consistency module for stereo generation; (3) we design a cross-domain self-distillation strategy and a perception-enhanced matcher that improve robustness on underwater degradations.


\section{Related Work}

\begin{figure*}[t]
  \centering
  \includegraphics[width=\textwidth]{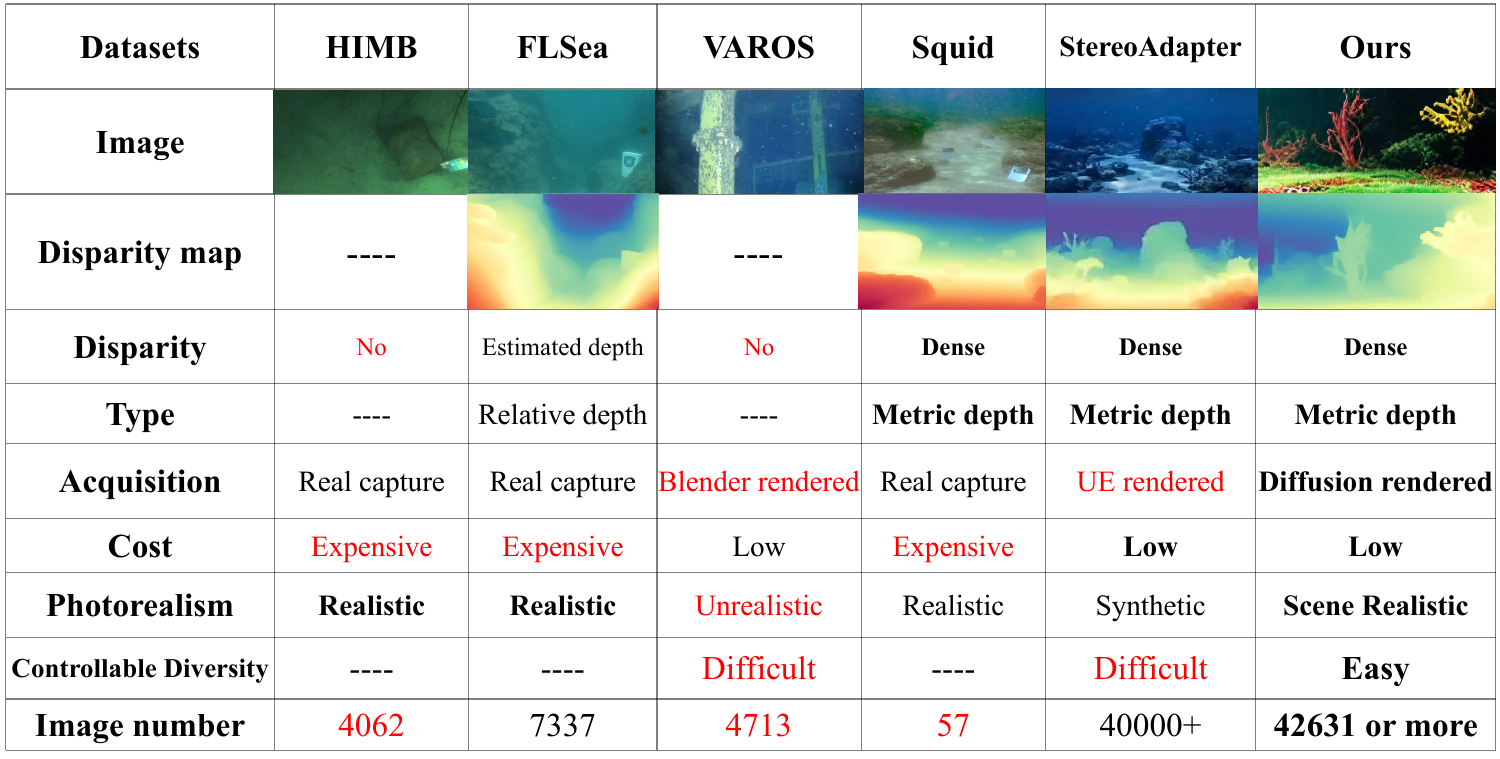}
\caption{\textbf{Statistics and comparison of existing underwater stereo datasets}. Including HIMB~\cite{Skinner2019UWstereoNet}, FLSea~\cite{Randall2023FLSea}, VAROS~\cite{Zwilgmeyer2021VAROS}, Squid~\cite{berman2020underwater}, StereoAdapter~\cite{wu2025stereoadapter} and our proposed dataset. The table summarizes key statistics such as the number of stereo pairs, capture conditions, and the availability of geometry annotations.}
  \label{fig:tab}
\end{figure*}

\noindent\textbf{Stereo Matching Datasets.}
Learning-based stereo matching primarily relies on annotated terrestrial datasets for supervised training, while underwater resources such as HIMB~\cite{Skinner2019UWstereoNet}, FLSea~\cite{Randall2023FLSea}, VAROS~\cite{Zwilgmeyer2021VAROS}, and UWStereo~\cite{UWStereo2024} remain limited in different ways, as summarized in Figure~\ref{fig:tab}. Some offer realistic imagery but lack dense ground truth depth or disparity (e.g., only relative depth or sparse labels). In contrast, others are synthetic with simplified optics, reduced photorealism, or restricted scene diversity and resolution, largely due to the cost and difficulty of underwater acquisition. To alleviate data scarcity, synthetic pipelines have been explored, including game-engine rendering, video forwarding to derive pseudo pairs, and 2D affine augmentations; although these strategies increase volume, they fail to reproduce underwater attenuation, backscatter, and turbidity, leaving a clear domain gap in real data. In contrast, our approach generates large-scale, diverse training sets from real-world stereo images by rendering geometry-preserving underwater counterparts, narrowing this gap without additional underwater collection.

\noindent\textbf{Image Editing Models.}
Latent diffusion models~\cite{Rombach2022LDM,zhang2026atlantis++,zhang2024atlantis,zou2025calibration} generate high-fidelity images in latent space and offer stronger stability and controllability than GAN-based stylization~\cite{goodfellow2014gan,dhariwal2021beatgans}, especially with classifier-free guidance~\cite{ho2022classifierfree} and ControlNet-style structural conditioning~\cite{zhang2023controlnet,saharia2022imagen}. Beyond image synthesis, diffusion priors have been applied to dense prediction tasks such as optical flow and monocular depth estimation~\cite{saxena2024surprising,ke2024repurposing,liang2025distilling,liang2025flow,Li2025CVMJ}. Recent diffusion-based monocular depth methods further synthesize challenging yet depth-consistent appearances from clean images, improving robustness under adverse conditions~\cite{tosi2024diffusion,wang2025robustereo,2024relation,11075607}. However, extending such image editing strategies from monocular depth to stereo matching is non-trivial: the generated left and right views must be realistic while preserving cross-view correspondence, epipolar consistency, and disparity scale. Independently editing the two views may introduce inconsistent textures, boundaries, color shifts, or scattering patterns, directly corrupting cost-volume matching.

\noindent\textbf{Stereo Matching Networks.}
Learning-based stereo matching has replaced traditional pipelines with convolutional neural networks (CNNs). GCNet~\cite{kendall2017gcnet} first regularized the 4D cost volume using 3D convolutions. Subsequent models such as PSMNet~\cite{chang2018psmnet}, GANet~\cite{zhang2019ganet} and GwcNet~\cite{guo2019gwcnet} improved accuracy via hierarchical aggregation, group-wise correlation, and guided aggregation, while the cascade-based CFNet~\cite{shen2021cfnet} further enhanced efficiency and generalization. Building on these advances, IGEV~\cite{xu2023iterative} introduced a geometry encoding volume that captures non-local context. The field has since shifted toward more generalizable architectures: IGEV++~\cite{xu2025igev++} advances geometry-aware encoding and iterative refinement, and vision foundation models~\cite{wang2025boosting} such as FoundationStereo~\cite{wen2025stereo} leverage large-scale pretraining to provide transferable backbones that set new baselines. However, most of these methods are developed for normal viewing conditions and degrade under underwater artifacts, making robustness in such challenging environments a key open problem.

\section{Method}

\begin{figure*}[t]
  \centering
  \includegraphics[width=\textwidth]{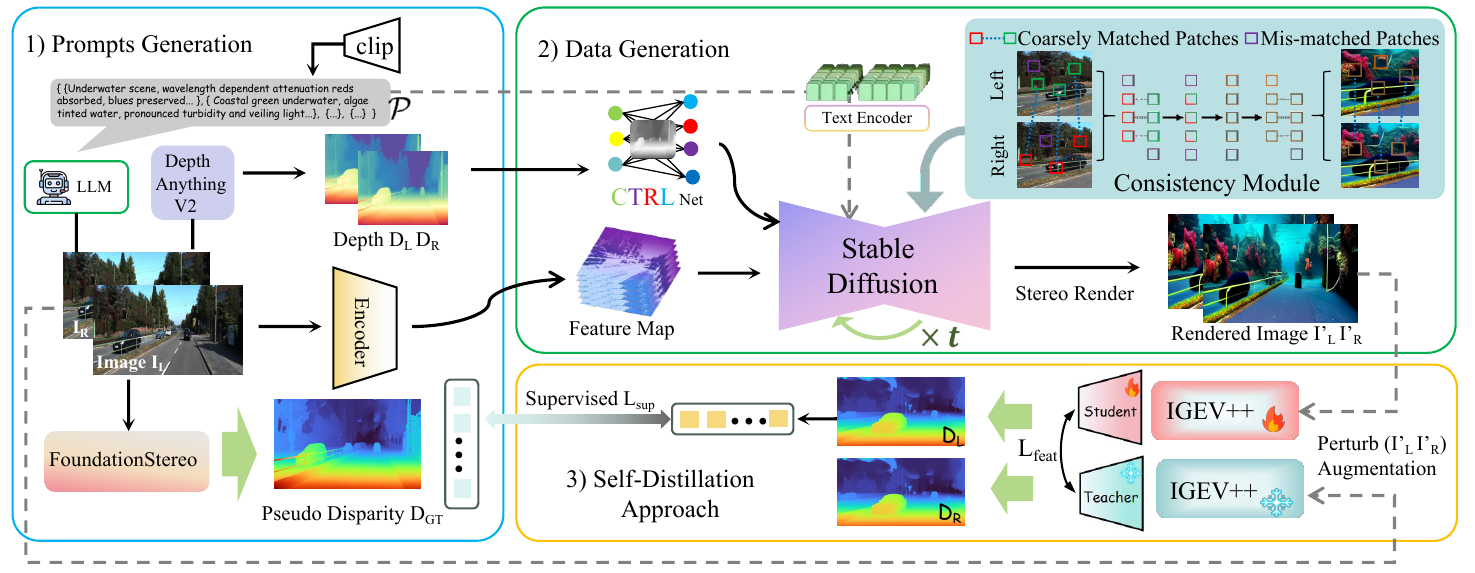}
\caption{\textbf{Overview of our framework.}
The pipeline comprises three parts:
(1) Prompts Generation: brief underwater descriptors (via
an LLM or CLIP~\cite{clip}) and a monocular depth map is extracted from a clean terrestrial pair to describe underwater scenes.
(2) Data Generation: the text prompt and depth map condition a ControlNet-guided diffusion model; a lightweight patch-level consistency module enforces left-right coherence and yields geometry-preserving underwater stereo images.
(3) Self-Distillation Approach: the frozen teacher operates on the clean pair, while the student ingests the rendered/perturbed underwater pair.}
  \label{fig:1}
\end{figure*}

In this section, we first present the problem formulation and motivation. We then describe the data generation pipeline with self-distillation, shown in Figure~\ref{fig:1}. Finally, we detail the improved stereo matching network with a perception-enhanced module, shown in Figure~\ref{fig:2}.

\subsection{Formulation and Motivation}

Accurate zero-shot stereo matching underwater remains challenging: attenuation, backscatter, turbidity, wavelength dependent color casts, and low illumination reduce contrast and texture, making correspondence unreliable (Figure~\ref{fig:yuanli}). A major bottleneck is data. Acquiring underwater stereo with accurate metric depth is costly and difficult---active sensors (e.g., underwater LiDAR/structured light) are expensive, range-limited, and often sparse, while stereo capture requires careful calibration under refraction and stable hardware in dynamic conditions. Consequently, real datasets (e.g., Sea-thru~\cite{Sea-Thru} and Squid~\cite{berman2020underwater}) remain small and supervision is limited. To mitigate data scarcity, prior work uses GAN-based~\cite{Hambarde2021UWGANSD,LI2020107038} translation and style transfer to make terrestrial images look ``underwater,'' but results are often stylized and still exhibit a clear domain gap.

This is where \textbf{AquaStereo} comes into play. We introduce a scalable depth-conditioned generation pipeline that synthesizes diverse underwater stereo pairs from depth maps and concise textual prompts, enabling virtually unlimited sampling while keeping geometric structure intact. Building on this data, we further employ cross-domain self-distillation to transfer reliable geometric cues from clean-domain stereo to underwater statistics. Finally, we augment the matcher with learnable perception frames that absorb underwater priors during training, improving feature extraction under severe degradations. We also incorporate a lightweight left-right consistency module to reduce stereo inconsistencies and improve geometric coherence.

\subsection{Stereo Data Generation for Underwater Scenes}
\label{sec:gen}

To address the scarcity of underwater stereo data, we build a scalable simulation pipeline that converts terrestrial stereo pairs into geometry-preserving underwater counterparts. Since dense, reliable disparity supervision is often sparse or noisy in real-world terrestrial datasets, we first collect a large corpus of high-quality rectified stereo pairs and compute pseudo disparities using a strong foundation stereo model (e.g., FoundationStereo~\cite{wen2025stereo}):
\begin{equation}
D_{\mathrm{GT}} = F_{FStereo}(I_L, I_R).
\end{equation}
Each sample $(I_L, I_R, D_{\mathrm{GT}})$ lies in the normal-domain dataset. Our goal is to map it to the underwater domain with an operator that changes appearance while preserving binocular geometry.

\noindent\textbf{Depth-conditioned diffusion generation.}
Underwater degradation depends strongly on scene depth. We synthesize underwater appearance by conditioning a diffusion model on per-pixel depth maps~\cite{Yang2024DepthAnythingV2} $(Depth_{L}, Depth_{R})$, enabling distance-dependent haze and scattering without simply copying terrestrial colors. 
As illustrated in Figure~\ref{fig:1} (prompt generation), we construct a diverse text prompt pool $\mathcal{P}$ by: (i) collecting underwater-scene descriptions and extracting representative prompts via CLIP~\cite{clip}; and (ii) using an LLM to summarize key principles of underwater image formation from physics-based literature into concise prompts (e.g., wavelength-dependent attenuation, backscatter, and veiling light~\cite{Sea-Thru,Derya}). 
During generation, we randomly sample a prompt $p \sim \mathcal{P}$ and feed the depth maps together with $p$ into a ControlNet to obtain the guidance features $c_t$ (Figure~\ref{fig:1}, data generation). Conditioned on $c_t$, we use Stable Diffusion to synthesize the underwater stereo pair $(I'_L, I'_R)$:
\begin{align}
c_t \quad&= F_{\text{CtrlNet}}(z_t, Depth_{L}, Depth_{R}, p),\\
(I'_L, I'_R) &= F_{\text{StableDiffusion}}(z_t, p \mid c_t).
\end{align}
This procedure yields diverse yet geometry-faithful underwater stereo pairs.

\noindent\textbf{Coherence-Enhanced Consistency.}
Unlike single-image generation, diffusion may introduce stereo pair inconsistencies. 
We mitigate this using a lightweight module that samples patch pairs along epipolar lines, scores them with appearance cues and coarse disparity priors, and applies a single self-attention pass over mixed positive and negative matches. 
Concretely, for a left patch feature $\mathbf{f}_i^L$ and candidate right features $\mathbf{f}_j^R$, we define a disparity-aware similarity:
\begin{equation}
s_{ij}
= \lambda \,\frac{\langle \mathbf{f}_i^L, \mathbf{f}_j^R\rangle}
        {\|\mathbf{f}_i^L\|_2 \,\|\mathbf{f}_j^R\|_2}
  + (1-\lambda)\exp\!\Big(-\tfrac{|d_i - d_j|}{\tau}\Big),
\label{eq:coherence-score}
\end{equation}
where $d_i$ and $d_j$ are coarse disparity priors, and $\lambda\in[0,1]$ balances appearance and geometric cues. 
The resulting scores are normalized into soft confidence masks, which modulate the ControlNet features and improve binocular coherence with negligible overhead. The geometry-consistent synthetic pairs form our underwater training dataset. We summarize our dataset alongside representative underwater benchmarks in Figure~\ref{fig:tab}. Compared with real capture (costly and hard to scale) and CG rendering (low cost but often less photorealistic and harder to diversify), our diffusion-based generation provides a low-cost and scalable alternative with scene-realistic appearance and controllable diversity.

\subsection{Self-Distillation with Geometry Alignment}
\label{sec:distill}
To more tightly couple terrestrial and underwater domains and extend effective supervision across both, we adopt cross-domain self-distillation.
For each geometry-sharing pair
$\mathbf{x}^{\mathrm{norm}}=(I_L, I_R, D_{GT})$ and
$\mathbf{x}^{\mathrm{underwater}}=(I'_L, I'_R, D_{GT})$,
we employ a teacher--student scheme tied together by the common target $D_{GT}$.
The teacher $T$ is trained on clean terrestrial pairs
and remains frozen throughout distillation.
To further enhance the student’s learning from the teacher model, increase the diversity of underwater appearances, and improve robustness to underwater domain shifts, we apply an additional augmentation to the underwater branch:
\begin{equation}
(\tilde{I}'_L,\tilde{I}'_R)=\mathrm{Perturb}(I'_L,I'_R),
\end{equation}
where $\mathrm{Perturb}(\cdot)$ includes turbidity/backscatter amplification, color shifts, compression noise, and mixed or mismatched patch augmentations.

\noindent\textbf{Self-Distillation Approach.}\;
As shown in Figure~\ref{fig:1} (Self-Distillation Approach), given the terrestrial pair $(I_L,I_R)$, the teacher extracts multi-scale features
$F_T$.
For each underwater counterpart $(\tilde{I}'_L,\tilde{I}'_R)$, the student extracts features
$F_S$.
We align geometry across domains by matching features at the feature-extraction stage. Concretely, we apply an $\ell_1$ feature alignment loss with a stage weight:
\begin{equation}
\mathcal{L}_{\mathrm{feat}}
= \sum_{k=1}^{K}
\big\|\, F_S(\tilde{I}'_L,\tilde{I}'_R)-F_T(I_L,I_R)\,\big\|_{1},
\label{eq:lfeat}
\end{equation}
where $F_T(\cdot)$ and $F_S(\cdot)$ denote the teacher and student feature extractors, respectively, and $K$ is the number of perturbed underwater counterparts sampled via $\mathrm{Perturb}(\cdot)$ for the same geometry-sharing pair. This feature alignment encourages the student to retain teacher-consistent geometric cues under underwater degradations without requiring any weight sharing.

\noindent\textbf{Supervised loss for scale stability.}\;
To prevent source-domain drift and preserve metric scale, we supervise the student on the unperturbed pair using the shared pseudo disparity $D_{\mathrm{GT}}$:
\begin{equation}
\mathcal{L}_{\mathrm{sup}}
= \sum_{k=1}^{K}
\big\|\, D_S(I_L',I_R') - D_{\mathrm{GT}} \,\big\|_{1},
\label{eq:sup}
\end{equation}
where $D_S(\cdot)$ denotes the student’s disparity prediction. This term stabilizes iterations and calibrates the student to the teacher before transferring to underwater statistics. In sum, the training objective includes the two sequence terms above: \begin{equation} \mathcal{L} = \mathcal{L}_{\mathrm{feat}} \;+\; \lambda_{\mathrm{sup}}\,\mathcal{L}_{\mathrm{sup}}, \label{eq:total} \end{equation} with no additional regularization. We train jointly on both domains, use identical parameters for the two branches, keep the teacher frozen and parameter-disjoint, and apply perturbations only to the underwater inputs so as to widen the non-semantic gap while still preserving geometry through the shared target $D_{GT}$.

\subsection{Perception-Enhanced Network}
\label{sec:net}

\noindent\textbf{Learning a perception-enhanced encoder.} Underwater stereo matching is highly sensitive to feature degradation. Attenuation, backscatter, turbidity, and low texture weaken local correspondences and make cost-volume construction ambiguous, especially around object boundaries and texture-poor regions. Standard image encoders usually extract features from the left and right views,
\begin{wrapfigure}{r}{0.48\columnwidth} 
\centering 
\captionsetup{hypcap=false} 
\includegraphics[width=\linewidth]{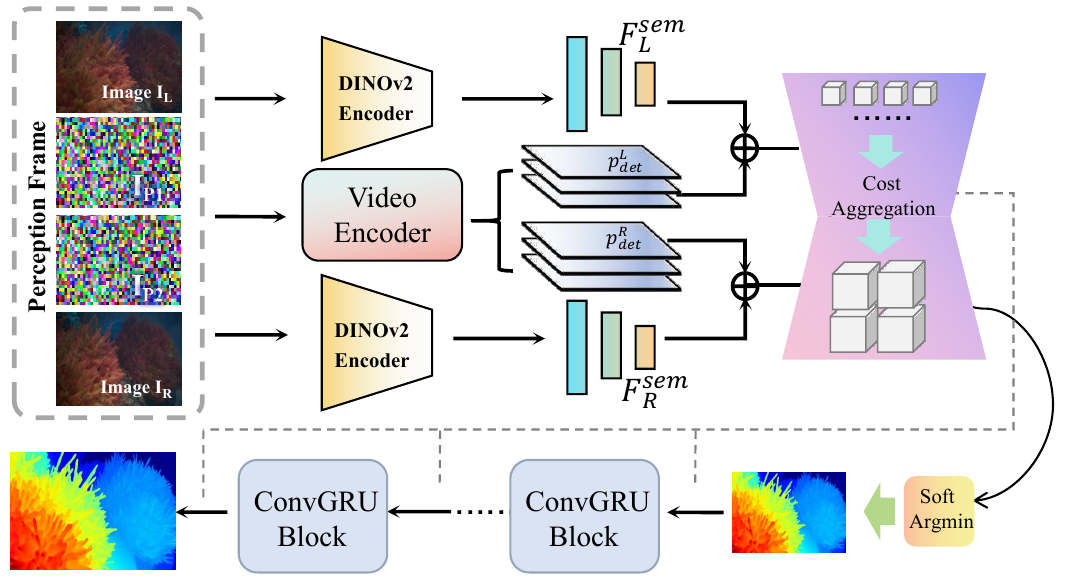} 
\caption{\textbf{Perception-enhanced stereo matcher.} We append two perception frames to the stereo pair, fuse the encoded features with DINOv2~\cite{oquab2023dinov2} semantics, and iteratively refine multi-range cost volumes to predict disparity.} 
\label{fig:2} 
\end{wrapfigure}
which limits their ability to model cross-view interactions before matching. Although a rectified stereo pair is not a temporal video, a video encoder provides a natural mechanism for cross-frame attention. We therefore formulate the stereo pair as a pseudo temporal cross-view sequence, where cross-frame attention serves as a proxy for cross-view feature alignment. To enrich this pseudo temporal context and capture underwater degradation priors, 
we augment a stereo pair $(I_L, I_R)$ with two learnable perception frames $I_{P1}$ 
and $I_{P2}$, as shown in Figure~\ref{fig:2}. The four frames are arranged in temporal 
order and passed through a video encoder to obtain multi-scale features:
\begin{equation}
\mathbf{f}_{left},\mathbf{f}_{right}\;=\;\mathcal{F}_{\mathrm{enc}}\!\bigl(I_{P1}\ \mathbin{\textcircled{c}}\ I_L\ \mathbin{\textcircled{c}}\ I_R\ \mathbin{\textcircled{c}}\ I_{P2}\bigr),
\label{eq:frencoder}
\end{equation}

where $\mathbin{\textcircled{c}}$ denotes temporal concatenation and $\mathcal{F}$ is a video encoder following~\cite{change3d}.
Let $f_{\text{left}}$ and $f_{\text{right}}$ denote the view-aligned features extracted from the left and right frames, respectively, via temporal modeling.
The perception frames are learnable parameters that act as degradation-aware visual prompts. During training, they absorb underwater priors and help the encoder produce more stable, view-consistent, and discriminative features under challenging degradations. In parallel, we extract high-level semantics from the raw stereo views using a DINOv2~\cite{oquab2023dinov2} image backbone:
$F^{\mathrm{sem}}_{L}=\mathcal{E}_{\mathrm{img}}(I_L)$ and $F^{\mathrm{sem}}_{R}=\mathcal{E}_{\mathrm{img}}(I_R)$,
where $\mathcal{E}_{\mathrm{img}}$ is a shared-weight feature extractor.
We then fuse the perception features with their semantic counterparts via channel concatenation followed by a $1{\times}1$ projection, yielding the left-right matching descriptors:
\begin{equation}
p_L \;=\; g\!\left([\,f_{\text{left}};\,F^{\mathrm{sem}}_{L}\,]\right), \quad
p_R \;=\; g\!\left([\,f_{\text{right}};\,F^{\mathrm{sem}}_{R}\,]\right).
\label{eq:dino-fuse}
\end{equation}
The fused features $(p_L,p_R)$ are used to build the cost volume and are fed into an iterative refinement with Soft-Argmin regression, following IGEV++~\cite{xu2025igev++}.

\begin{table*}[t]
\caption{\textbf{Comparison with state-of-the-art stereo methods on the UWStereo~\cite{UWStereo2024} benchmark.}
"Ours" is trained on the rendered underwater dataset, whereas most baselines are trained on SceneFlow~\cite{Mayer2016SceneFlow}, except LightStereo~\cite{guo2025lightstereo}, StereoAnything~\cite{guo2024stereo}, NMRF~\cite{guan2024nmrfstereo}, MonSter~\cite{monster}, and FoundationStereo~\cite{wen2025stereo}, which use mixed synthetic and real data.
}

\label{tab:underwater_benchmark}
\centering
\scriptsize 
\begin{tabular}{l*{10}{c}}
\toprule
\multirow{2}{*}{Method} &
\multicolumn{2}{c}{Coral} & \multicolumn{2}{c}{Default} &
\multicolumn{2}{c}{Industry} & \multicolumn{2}{c}{Ship} &
\multicolumn{2}{c}{Total} \\
\cmidrule(lr){2-3}\cmidrule(lr){4-5}\cmidrule(lr){6-7}\cmidrule(lr){8-9}\cmidrule(lr){10-11}
 &  EPE$\downarrow$ & D1$\downarrow$ & EPE$\downarrow$ & D1$\downarrow$
 & EPE$\downarrow$ & D1$\downarrow$ & EPE$\downarrow$ & D1$\downarrow$
 & EPE$\downarrow$ & D1$\downarrow$ \\
\midrule
PSMNet~\cite{chang2018psmnet} \venue{[CVPR'18]}                 & 2.680 & 13.170 & 2.360 & 7.210  & 2.850 & 10.580 & 4.950 & 15.700 & 3.540 & 12.360 \\
CFNet~\cite{shen2021cfnet} \venue{[CVPR'21]}                   & 2.503 & 10.043 & 2.060 & 7.615  & 2.605 & 10.518 & 4.903 & 16.123 & 3.368 & 12.092 \\
GwcNet~\cite{guo2019gwcnet} \venue{[CVPR'19]}                  & 2.600 & 10.668 & 4.162 & 9.750  & 4.101 & 13.367 & 7.438 & 20.295 & 5.138 & 14.986 \\
IGEV~\cite{xu2023iterative} \venue{[CVPR'23]}                  & 2.262 & 12.310 & 1.088 & 5.580  & 1.990 & 7.590  & 2.654 & 12.300 & 2.138 & 9.758  \\
LightStereo~\cite{guo2025lightstereo} \venue{[ICRA'25]}        & 2.503 & 11.209 & 1.776 & 7.542  & 2.551 & 10.598 & 3.966 & 16.668 & 2.952 & 12.487 \\
StereoAnything~\cite{guo2024stereo} \venue{[ICCV'23]}   & 2.117 & 8.518  & 1.118 & 5.539  & 1.527 & 8.058  & 3.087 & 13.187 & 2.138 & 9.655  \\
NMRF~\cite{guan2024nmrfstereo} \venue{[CVPR'24]}               & 3.036 & 14.580 & 2.734 & 10.570 & 2.676 & 12.610 & 5.353 & 18.390 & 3.747 & 14.758 \\
MonSter~\cite{monster} \venue{[CVPR'25]}                       & 1.882 & 9.010  & \second{0.664} & \second{2.940} & 1.061 & 4.870 & \second{1.254} & \second{6.340} & \second{1.195} & \second{5.744} \\
FoundationStereo~\cite{wen2025stereo}                          & \second{1.811} & \second{7.590} & 0.844 & 3.090 & \second{0.718} & \second{2.490} & 2.750 & 8.890 & 1.669 & 5.768 \\
\midrule
\rowcolor{gray!15}
\textbf{AquaStereo (Ours)}                                   & \textbf{1.528} & \textbf{7.180} &
\textbf{0.338} & \textbf{1.320} &
\textbf{0.373} & \textbf{1.510} &
\textbf{0.500} & \textbf{2.240} &
\textbf{0.590} & \textbf{2.616} \\
\bottomrule
\end{tabular}

\end{table*}

\begin{figure}[t]
  \centering
  \includegraphics[width=\textwidth,height=81mm]{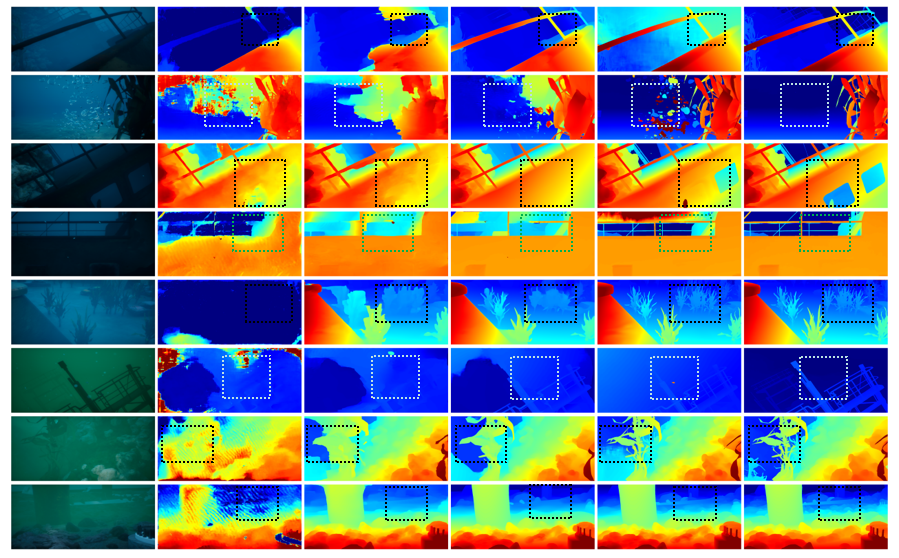}

\begin{minipage}{\textwidth}
    \centering
    \scriptsize
    \begin{tabular}{@{}*{6}{>{\centering\arraybackslash}p{0.16\textwidth}}@{}}
        Left Image
        & StereoBase
        & \shortstack{IGEV-KITTI}
        & IGEV-SF
        & \shortstack{FoundationStereo}
        & Ours
    \end{tabular}
\end{minipage}

\caption{\textbf{Qualitative results in underwater scenes.}
Our method produces more coherent disparity maps under underwater degradations, preserving sharper boundaries and finer structures with fewer artifacts.}

  \label{fig:contrast}
\end{figure}

\section{Experiments}
\label{sec:exp}

In this section, we first introduce the datasets and evaluation metrics for experiments. Then, detailed comparisons
are conducted with the state-of-the-art methods. Finally, ablations and discussions are performed to confirm the effectiveness of the main components.

\subsection{Experimental Setup}
\noindent\textbf{Datasets.}
We use terrestrial stereo datasets for supervised training and underwater benchmarks for evaluation. For training, SceneFlow~\cite{Mayer2016SceneFlow}, KITTI 2012~\cite{geiger2012kitti}, and KITTI 2015~\cite{Menze2015KITTI15} provide clean stereo pairs with reliable geometry to drive our rendering and learning pipeline. For underwater evaluation, we report results on FLSea\text{-}Stereo~\cite{Randall2023FLSea} and Squid~\cite{berman2020underwater} as real in-situ, forward\text{-}looking benchmarks captured under diverse underwater conditions. We additionally evaluate on UWStereo~\cite{UWStereo2024} as a physics\text{-}based synthetic benchmark with dense metric ground truth, and on the underwater split of TartanAir~\cite{tartanair2020iros} to assess generalization to simulated underwater scenes.

\noindent\textbf{Stereo matching models.}
We compare our approach with representative stereo matchers: PSMNet~\cite{chang2018psmnet}, CFNet~\cite{shen2021cfnet}, GwcNet~\cite{guo2019gwcnet}, IGEV~\cite{xu2023iterative}, IGEV++~\cite{xu2025igev++}, LightStereo~\cite{guo2025lightstereo}, StereoAnything~\cite{guo2024stereo}, NMRF~\cite{guan2024nmrfstereo}, MonSter~\cite{monster}, and FoundationStereo~\cite{wen2025stereo}. For the perception-enhanced encoder, we additionally use a Change3D-style video encoder~\cite{change3d} and DINOv2~\cite{oquab2023dinov2}. For self-distillation, both the teacher and student adopt the same IGEV++ architecture; the teacher is pre-trained on our terrestrial dataset using pseudo-disparities generated by FoundationStereo, and is then used to supervise the student during training. Unless otherwise stated, we follow official implementations and default settings.

\noindent\textbf{Stable Diffusion models.}
We implement our generation pipeline with the Diffusers library, adapting Stable Diffusion (SD) and ControlNet. We use SD v1.5 as the image generator with a DDIM scheduler and 50 sampling steps. For depth-conditioned generation, we employ ControlNet (control\_v11f1p\_sd15\_depth) together with Depth-Anything V2 for monocular depth estimation. Source images from SceneFlow and KITTI are used to synthesize stereo data in underwater conditions. We refer to the resulting synthetic training set as UW-Dataset (40K).

\noindent\textbf{Training Parameters.} Unless otherwise stated, we train with a learning rate of $10^{-4}$ (batch size $=8$ per GPU). Models are optimized at a resolution of $384\times736$ pixels, with the maximum disparity kept fixed across methods. Training takes approximately three days on 4 RTX 4090 (24GB) GPUs. For self-distillation, we adopt the frozen-teacher and student losses described in Sec.~\ref{sec:distill}, use the same model configuration, and set $\lambda_{\mathrm{sup}}=0.2$.

\noindent\textbf{Evaluation Metrics.}
Zero-shot generalization is evaluated on UWStereo~\cite{UWStereo2024}, FLSea~\cite{Randall2023FLSea}, Squid~\cite{berman2020underwater} and TartanAir~\cite{tartanair2020iros}.
We report EPE and the D1 outlier rate as our primary evaluation metrics.
EPE is defined as the mean absolute disparity error (in pixels) over all valid pixels.
D1 is the percentage of pixels with error larger than $\max(3\text{ px},\,0.05\, D)$, which serves as a stringent indicator of robustness and stability in challenging underwater scenes.

\begin{figure}[t]
\centering

\begin{minipage}[t]{0.48\textwidth}
    \vspace{0pt}
    \centering
    \captionsetup{type=table}
    \caption{\textbf{Comparison on UWStereo with different training sets.}
    We evaluate three representative stereo networks trained on SceneFlow~\cite{Mayer2016SceneFlow}, KITTI~\cite{Menze2015KITTI15}, or our rendered UW-Dataset.}
    \label{tab:uwstereo-3datasets-total}

    \fontsize{6}{8}\selectfont
    \begin{tabular}{llcc}
    \toprule
    \multirow{2}{*}{Datasets} & \multirow{2}{*}{Networks} &
    \multicolumn{2}{c}{Total} \\
    \cmidrule(lr){3-4}
     &  & EPE$\downarrow$ & D1$\downarrow$ \\
    \midrule
    SceneFlow~\cite{Mayer2016SceneFlow}
    & 
        & \second{3.5400} & \second{12.360} \\
    KITTI~\cite{Menze2015KITTI15}
    & PSMNet~\cite{chang2018psmnet}
        & 4.7630 & 17.558 \\
    \rowcolor{gray!15}\cellcolor{white}
    UW-Dataset
    & \cellcolor{white}
        & \best{1.1422} & \best{10.16} \\
    \midrule
    SceneFlow~\cite{Mayer2016SceneFlow}
    & 
        & \second{2.5930} & \second{10.861} \\
    KITTI~\cite{Menze2015KITTI15}
    & StereoBase~\cite{guo2023openstereo}
        & 3.8360 & 14.755 \\
    \rowcolor{gray!15}\cellcolor{white}
    UW-Dataset
    & \cellcolor{white}
        & \best{2.5070} & \best{10.402} \\
    \midrule
    SceneFlow~\cite{Mayer2016SceneFlow}
    & 
        & \second{2.1380} & \second{9.758} \\
    KITTI~\cite{Menze2015KITTI15}
    & IGEV++~\cite{xu2025igev++}
        & 4.0100 & 16.106 \\
    \rowcolor{gray!15}\cellcolor{white}
    UW-Dataset
    & \cellcolor{white} 
        & \best{1.7280} & \best{8.418} \\
    \bottomrule
    \end{tabular}
\end{minipage}
\hfill
\begin{minipage}[t]{0.48\textwidth}
    \vspace{0pt}
    \centering
    \includegraphics[width=\linewidth]{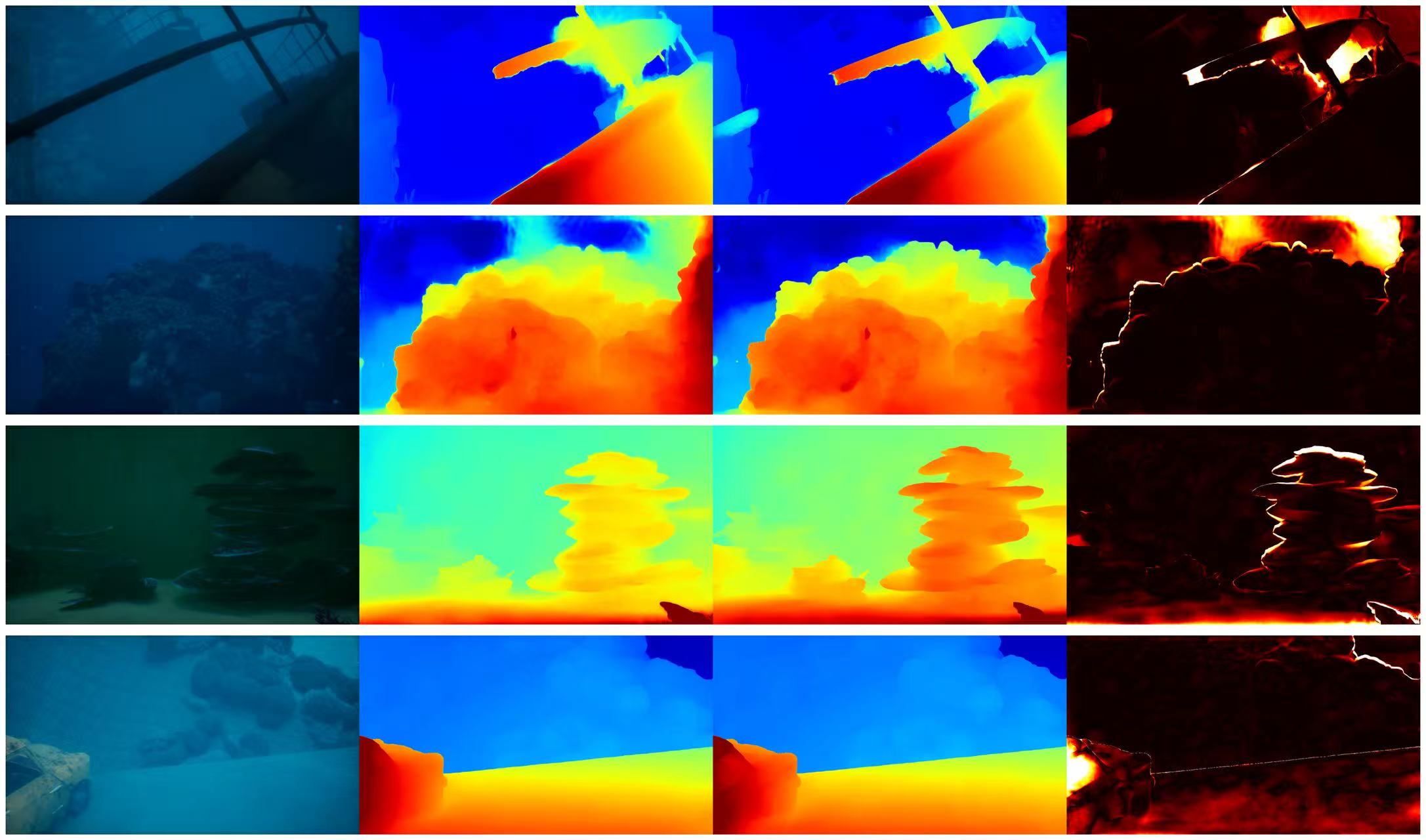}

    \vspace{-1mm}
    \begin{minipage}{\linewidth}
        \centering
        \tiny
        \begin{tabular}{@{}*{4}{>{\centering\arraybackslash}p{0.24\linewidth}}@{}}
            Input
            & \shortstack{Before Self-\\distillation}
            & \shortstack{After Self-\\distillation}
            & \shortstack{Heatmap of\\Difference}
        \end{tabular}
    \end{minipage}

    \captionsetup{type=figure}
\caption{\textbf{Effect of self-distillation on disparity maps.}
Self-distillation produces smoother and more complete disparities, while the heatmap highlights corrections around structural boundaries and degraded regions.}

    \label{fig:self-distill}
\end{minipage}

\end{figure}

\subsection{Quantitative Results}

\noindent\textbf{Comparison with stereo matching methods.} Table~\ref{tab:underwater_benchmark} compares stereo matching methods on UWStereo with four scene splits: Coral, Default, Industry, and Ship. Our model ranks first on all splits and overall, achieving the lowest Total EPE$\downarrow$/D1$\downarrow$ of 0.59/2.61. Notably, methods that rely heavily on photometric or brightness-constancy assumptions, such as NMRF~\cite{guan2024nmrfstereo}, degrade underwater, where absorption, scattering, and backscatter violate these priors. Moreover, although many baselines are trained on mixed datasets, our approach remains strong, highlighting the effectiveness of our pipeline, which combines depth-conditioned synthesis, a lightweight left--right consistency check, dual-domain self-distillation, and a perception-enhanced stereo framework.

\noindent\textbf{Comparison with other training sets.}
Table~\ref{tab:uwstereo-3datasets-total} and Table~\ref{tab:muti_datasets} compare models trained on SceneFlow~\cite{Mayer2016SceneFlow}, KITTI~\cite{Menze2015KITTI15}, and our rendered UW-Dataset and evaluated zero-shot on the four UWStereo splits; training on UW-Dataset yields the best total accuracy for all three representative architectures (PSMNet~\cite{chang2018psmnet}, StereoBase~\cite{guo2023openstereo}, IGEV++~\cite{xu2025igev++}) with consistent gains, and IGEV++ represents the strongest model. More detailed results for all tables are provided in the supplementary material. Note that models trained only on terrestrial data (SceneFlow/KITTI) degrade notably underwater, reflecting the domain gap caused by absorption, scattering, and backscatter. Also, by fixing the architecture to IGEV++ and varying only the training data, we compare several training sets of comparable scale, including SceneFlow, SceneFlow+KITTI, StereoAdapter~\cite{wu2025stereoadapter}, and our UW-Dataset (Table~\ref{tab:muti_datasets}). This indicates that training on terrestrial data alone (SceneFlow or SceneFlow+KITTI) generalizes poorly to underwater benchmarks. StereoAdapter yields only limited gains on real underwater scenes, likely because CG rendering cannot fully reproduce real underwater image formation. In contrast, our UW-Dataset consistently achieves better performance across FLSea~\cite{Randall2023FLSea}, Squid~\cite{berman2020underwater}, and TartanAir~\cite{tartanair2020iros}, suggesting that our diffusion-based generation produces more realistic underwater appearances than CG rendering and leads to more effective training.


\begin{figure*}[t]
  \centering
  \includegraphics[width=\textwidth]{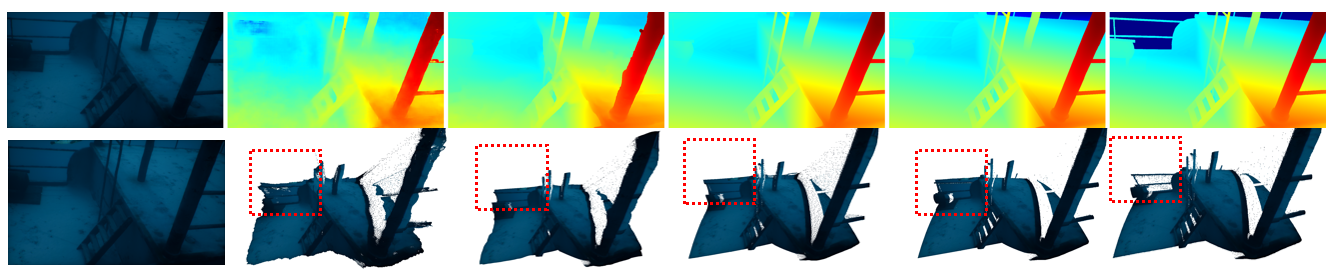}

\begin{minipage}{\textwidth}
    \centering
    \scriptsize
    \begin{tabular}{@{}*{6}{>{\centering\arraybackslash}p{0.16\textwidth}}@{}}
        Left Image
        & StereoBase
        & \shortstack{IGEV-KITTI}
        & IGEV-SF
        & \shortstack{FoundationStereo}
        & Ours
    \end{tabular}
\end{minipage}

\caption{\textbf{Qualitative point-cloud reconstructions in underwater scenes.}
Compared with multiple baselines, our method produces denser and cleaner reconstructions with fewer holes, floating outliers, and depth artifacts.}

  \label{fig:5}
\end{figure*}

\begin{table}[t]
\centering
\caption{\textbf{Comparison on underwater benchmarks with different training sets.}
We evaluate IGEV++~\cite{xu2025igev++} trained on SceneFlow~\cite{Mayer2016SceneFlow}, KITTI~\cite{Menze2015KITTI15}, UWStereo~\cite{UWStereo2024}, StereoAdapter~\cite{wu2025stereoadapter} or our rendered UW-Dataset on three underwater datasets: FLSea (Canyon/Rock)~\cite{Randall2023FLSea}, Squid~\cite{berman2020underwater}, and TartanAir~\cite{tartanair2020iros}.}
\label{tab:muti_datasets}
{\scriptsize
\begin{tabular}{ll*{8}{c}}
\toprule
\multirow{2}{*}{Datasets} & \multirow{2}{*}{Networks} &
\multicolumn{2}{c}{FL-Canyon~\cite{Randall2023FLSea}} & \multicolumn{2}{c}{FL-Rock~\cite{Randall2023FLSea}} &
\multicolumn{2}{c}{Squid~\cite{berman2020underwater}} & \multicolumn{2}{c}{TartanAir~\cite{tartanair2020iros}} \\
\cmidrule(lr){3-4}\cmidrule(lr){5-6}\cmidrule(lr){7-8}\cmidrule(lr){9-10}
 &  & EPE$\downarrow$ & D1$\downarrow$
    & EPE$\downarrow$ & D1$\downarrow$
    & EPE$\downarrow$ & D1$\downarrow$
    & EPE$\downarrow$ & D1$\downarrow$ \\
\midrule
SceneFlow~\cite{Mayer2016SceneFlow}
& \multirow{4}{*}{}
    & 5.41  & 35.82
    & 5.09  & 22.62
    & 2.18  & 12.28
    & 0.62  & 7.81 \\

SF~\cite{Mayer2016SceneFlow}+KITTI~\cite{Menze2015KITTI15}
& 
    & 5.10  & 22.76
    & 5.20  & 21.82
    & 2.02  & 11.55
    & 0.60  & 7.43 \\
    
UWStereo~\cite{UWStereo2024}
& IGEV++~\cite{xu2025igev++}
    & 5.01  & 21.45
    & 4.96  & 20.73
    & 2.08  & 12.64
    & 0.63  & 7.85\\

StereoAdapter~\cite{wu2025stereoadapter}
& 
    & \second{4.05} & \second{20.51}
    & \second{4.15} & \second{19.58}
    & \second{1.86} & \second{10.98}
    & \best{0.52} & \best{6.81} \\

\rowcolor{gray!15}\cellcolor{white}
UW-Dataset (Ours)
& \cellcolor{white}
    & \best{3.05} & \best{18.31}
    & \best{3.13} & \best{17.44}
    & \best{1.80}   & \best{10.52}
    & \second{0.55} & \second{6.92} \\
\bottomrule
\end{tabular}
}

\end{table}

\noindent\textbf{Qualitative Results.}
As shown in Figure~\ref{fig:contrast}, the disparity visualizations highlight the robustness of our method under strong color cast, depth-dependent attenuation, and heavy backscatter. In challenging regions such as the ship structure and surrounding rigging, our predictions maintain crisp object boundaries and smooth planar gradients, whereas baselines often exhibit bleeding artifacts and fragmented surfaces. Notably, transparent or semi-transparent phenomena (e.g., bubble-like structures and waterborne particles) frequently induce spurious matches in prior methods; in contrast, our results suppress these outliers and yield more coherent geometry. Thin structures and vegetation also remain continuous across the scene, preserving fine details in near- and mid-range areas that are critical for reliable underwater perception.

Moreover, Figure~\ref{fig:5} shows that these improvements translate into higher-quality 3D reconstructions. The reconstructed point clouds from our disparities are denser and more complete, with fewer floating outliers and reduced depth discontinuity noise. Fine obstacle geometry becomes clearly visible: for example, fence-like structures and slender bars are recovered with consistent depth and sharp contours. Such obstacles can be missed when disparities are noisy or over-smoothed, potentially compromising downstream tasks (e.g., underwater robot navigation and inspection) where failing to detect barriers may lead to unsafe trajectories or collisions. We also observe more stable seafloor topology and more complete hull surfaces, indicating that improvements are not limited to a single object type but generalize across both textured and low-texture regions. Overall, the point-cloud evidence corroborates that our more accurate disparities enable faithful, structure-preserving underwater 3D reconstruction.

\subsection{Ablation Study}

\noindent\textbf{Coherence-Enhanced Consistency Module.}
In this ablation, we isolate the effect of left-right (LR) consistency module under the same data-generation setting. As shown in Table~\ref{tab:r3_lr_consistency}, disabling the module during generation degrades the downstream stereo training results. Without enforcing cross-view coherence, the generated stereo pairs exhibit more inconsistencies, weakening correspondence cues and making matching ambiguous. Enabling our LR-consistency module reduces the inconsistency rate (Break px) from 15 to 5 with comparable generation cost, leading to improved EPE and D1.

\begin{table}[t]
\centering
\scriptsize

\caption{\textbf{LR-consistency ablation on UWStereo~\cite{UWStereo2024}}.
We evaluate LR consistency in terms of disparity quality and generation efficiency.
Gen. Time is the average time per image, and Break px denotes LR-inconsistent pixels per 100 pixels.}

\label{tab:r3_lr_consistency}
\begin{tabular}{l|l|cc|c|c}
\hline
\textbf{Module} & \textbf{Methods} & \textbf{EPE$\downarrow$} & \textbf{D1$\downarrow$}  & \textbf{Gen. Time (s) }$\downarrow$ & \textbf{Break px}$\downarrow$\\
\hline
 & Off & 3.237 & 14.82 & 8.4 & 15 \\
 \rowcolor{gray!15} \cellcolor{white}\multirow{-2}{*}{LR-Consistency Module} & On  & \textbf{2.953} & \textbf{12.14} & \textbf{8.1} & \textbf{5} \\
\hline

\end{tabular}

\end{table}

\begin{table*}[t]
\centering

\begin{minipage}[t]{0.48\textwidth}
\centering
\caption{\textbf{Ablation of self-distillation variants on the UWStereo benchmark~\cite{UWStereo2024}.}
All methods use the same IGEV++~\cite{xu2025igev++} stereo backbone. We compare a baseline without self-distill to self-distill variants including without weight sharing, shared extractor, shared iterative refinement, shared cost aggregation, L2 regularization, and plus supervised.}
\label{tab:ablation-distill-total}
\scriptsize
\begin{tabular}{l*{2}{c}}
\toprule
\multirow{2}{*}{Distillation variant} &
\multicolumn{2}{c}{Total} \\
\cmidrule(lr){2-3}
 & EPE$\downarrow$ & D1$\downarrow$ \\
\midrule
Baseline without self-distill             & 2.9530          & 12.144 \\
Self-distill without weight sharing       & 2.7510          & 11.195 \\
Self-distill + Share extractor    & 2.7740          & 11.262 \\
Self-distill + Share iterative         & 2.8030          & 11.265 \\
Self-distill + Share aggregation          & \second{2.7200} & \second{11.120} \\
Self-distill + L2 regularization          & 3.0510          & 13.838 \\
\rowcolor{gray!12}
\textbf{Self-distill + supervised}        & \best{2.6350}   & \best{10.549} \\
\bottomrule
\end{tabular}
\end{minipage}
\hfill
\begin{minipage}[t]{0.48\textwidth}
\centering
\captionof{table}{\textbf{Ablation of feature designs on the UWStereo benchmark~\cite{UWStereo2024}.} All variants share the same IGEV++~\cite{xu2025igev++} stereo matcher and differ only in the feature extractor, including IGEV++ backbone, VGG19~\cite{simonyan2015vgg}+DINOv2~\cite{oquab2023dinov2}, FoundationStereo features~\cite{wen2025stereo}, VGGT-style attention block~\cite{wang2025vggt}, ResNet~\cite{he2016resnet}+DINOv2~\cite{oquab2023dinov2}, and AquaStereo (ours).}
\label{tab:ablation-backbones-total}
\scriptsize
\begin{tabular}{l*{2}{c}}
\toprule
\multirow{2}{*}{Feature extractor variant} &
\multicolumn{2}{c}{Total} \\
\cmidrule(lr){2-3}
 & EPE$\downarrow$ & D1$\downarrow$ \\
\midrule
IGEV++ backbone                  & \second{2.9530} & \second{12.144} \\
VGG19 + DINOv2 & 3.0790 & 12.955 \\
FoundationStereo features      & 3.9290 & 15.429 \\
VGGT-style attention             & 3.4090 & 15.462 \\
ResNet + DINOv2    & 3.1410 & 13.197 \\
\rowcolor{gray!12}
\textbf{AquaStereo (Ours)}                             & \best{2.4830} & \best{10.824} \\
\bottomrule
\end{tabular}
\end{minipage}

\end{table*}

\noindent\textbf{Effectiveness of the self-distillation method.}
A binocular stereo matcher typically follows a three-stage pipeline: feature extraction, cost aggregation, and iterative refinement, as illustrated in Figure~\ref{fig:2}. Under identical training conditions, we evaluated several teacher--student coupling strategies by sharing weights at different stages of this pipeline. As reported in Table~\ref{tab:ablation-distill-total}, parameter-disjoint self-distillation already improves robustness over the non-distilled baseline. Sharing the feature extractor, the iterative refinement, or the aggregation module between teacher and student yields comparable performance and only modest additional gains, suggesting that the benefit mainly comes from cross-domain supervision rather than architectural coupling. In contrast, adding $\ell_2$ regularization restricts adaptation and degrades accuracy. The strongest performance is achieved when underwater-branch distillation is coupled with supervised learning on the clean branch using the shared pseudo target $D_{\mathrm{GT}}$, underscoring the value of geometry-aligned cross-domain training.

\noindent\textbf{Effectiveness of the perception-enhanced encoder.}
Table~\ref{tab:ablation-backbones-total} compares feature extractor designs under the same training protocol and IGEV++ stereo head, so performance differences can be attributed to the encoder. We evaluate several alternatives, including CNN backbones (VGG19~\cite{simonyan2015vgg}/ResNet~\cite{he2016resnet}) augmented with DINOv2 semantics~\cite{oquab2023dinov2}, a VGGT-style attention block~\cite{wang2025vggt}, and a FoundationStereo-like extractor~\cite{wen2025stereo}. While these variants improve representation capacity, they still struggle to produce stable correspondences under underwater degradations, where contrast loss, backscatter, and color casts weaken texture cues and amplify ambiguous matches. In contrast, \textbf{AquaStereo} achieves the best Total EPE/D1 (2.483/10.824), outperforming the IGEV++ backbone (2.953/12.144) and all other designs. We attribute the gains to two factors: learnable perception frames provide view-aligned, degradation-adaptive cues, while DINOv2 contributes high-level semantics that remain reliable when local texture is corrupted. Their fusion yields more discriminative matching descriptors, reducing cost-volume ambiguity and improving final disparities.

\noindent\textbf{Visualizing self-distillation with heatmaps.}
As shown in Figure~\ref{fig:self-distill}, we compare predictions before and after self-distillation and visualize their differences with heatmaps. The distilled model yields more coherent depth estimates, with sharper object boundaries, better-preserved thin structures, and fewer local artifacts such as flying pixels and halo effects. It also improves large planar regions by reducing depth bleeding and fragmentation in low-texture underwater areas. The heatmaps reveal that the main corrections are concentrated around depth discontinuities, specular regions, and haze-degraded areas, where stereo correspondence is typically ambiguous. These observations suggest that self-distillation enhances both local detail recovery and global depth consistency, leading to more reliable underwater geometry.

\begin{figure*}[t]
    \centering
    \scriptsize
    \captionsetup[subfloat]{labelformat=empty}

    \subfloat[\small Image]{
        \includegraphics[width=0.23\linewidth,height=0.14\linewidth]{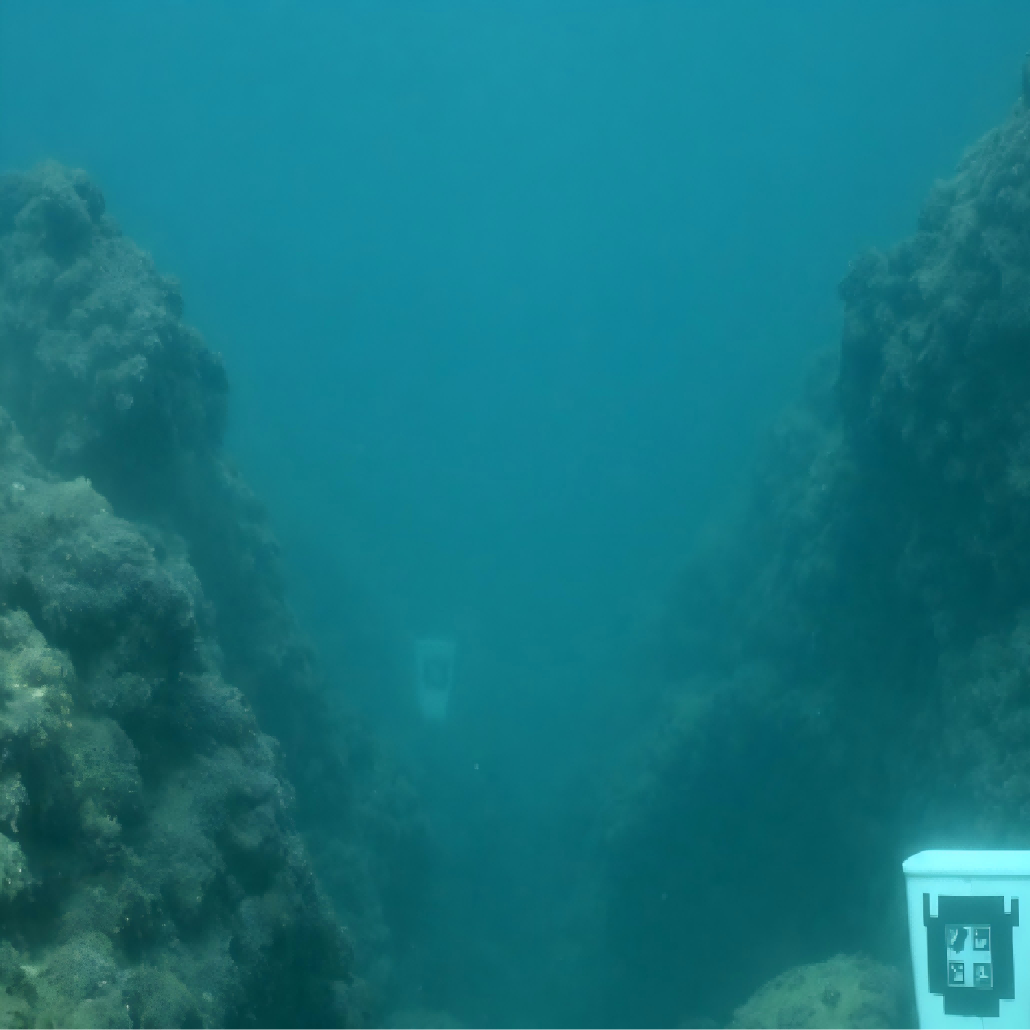}
    }%
    \subfloat[\small attention map]{
        \includegraphics[width=0.23\linewidth,height=0.14\linewidth]{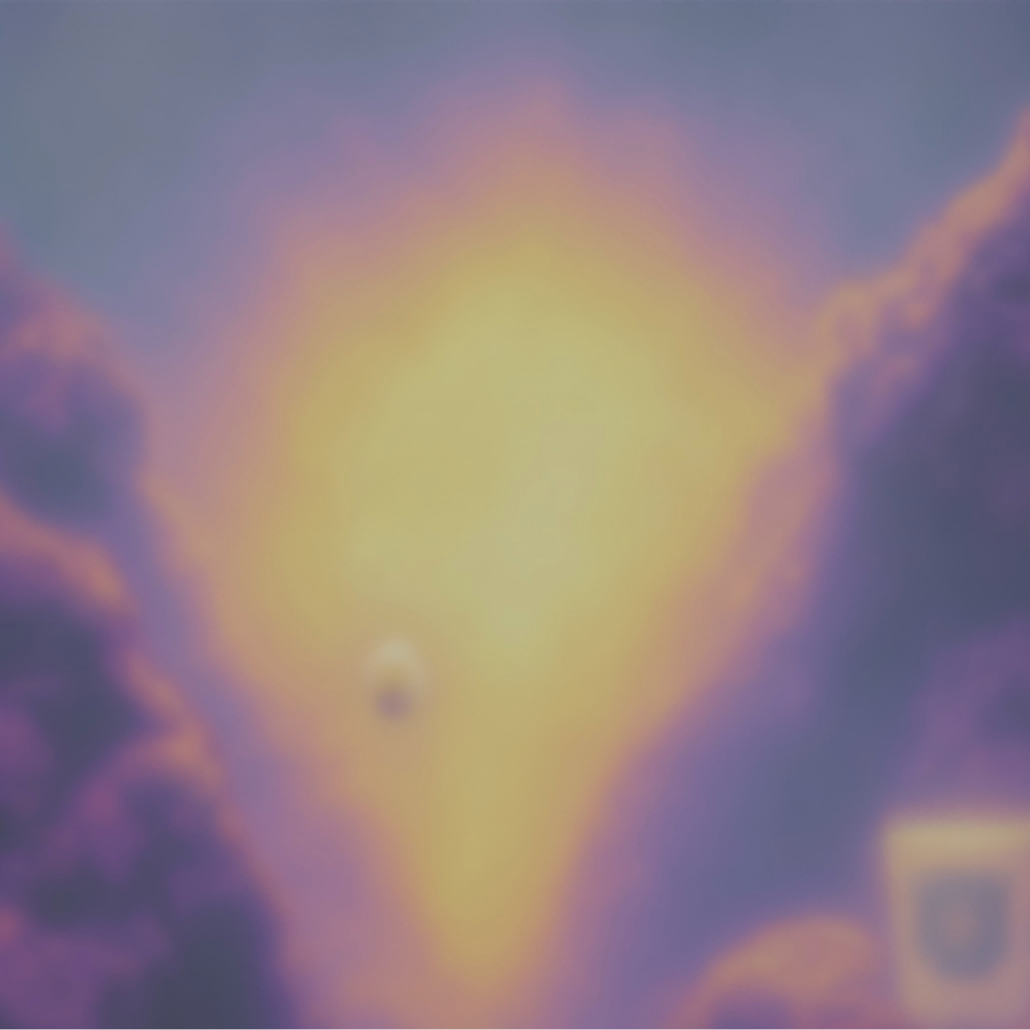}
    }%
    \subfloat[\scriptsize w/o perception frame]{
        \includegraphics[width=0.23\linewidth,height=0.14\linewidth]{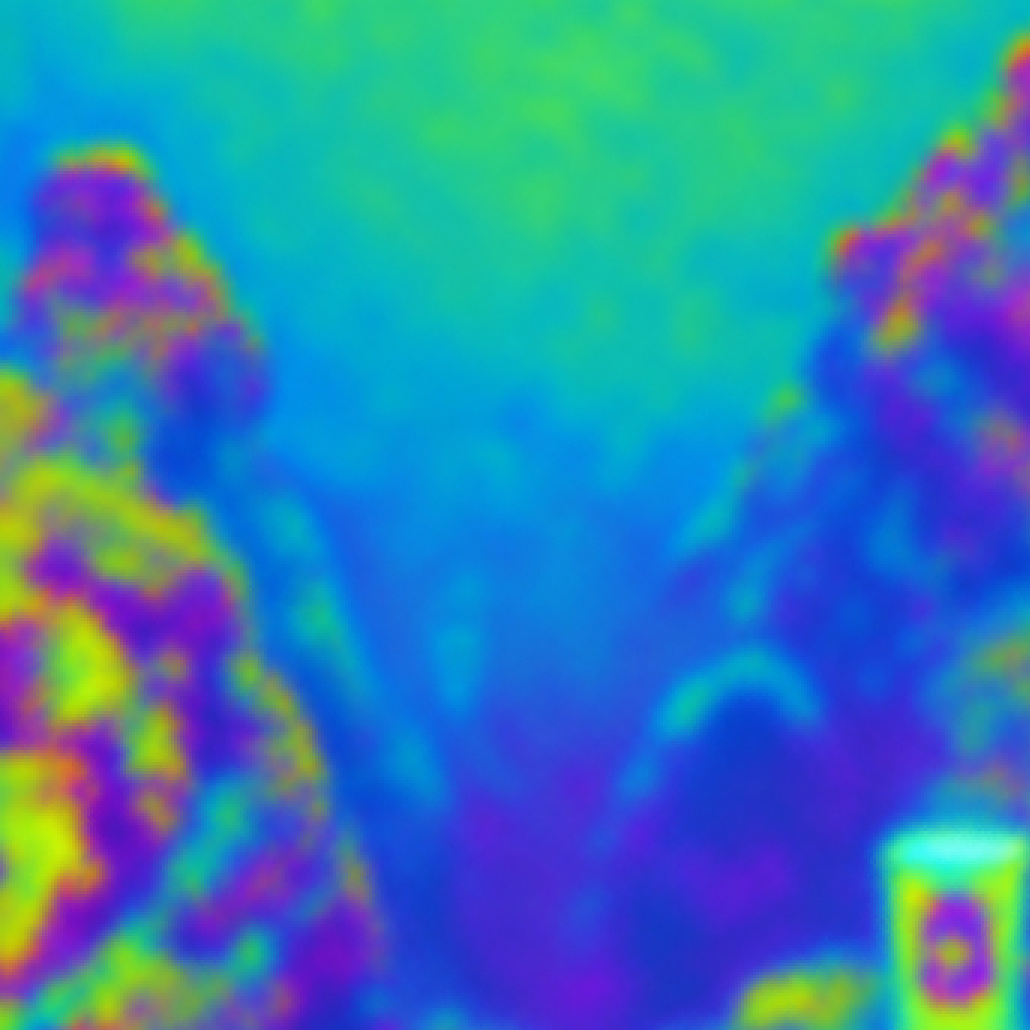}
    }%
    \subfloat[\scriptsize w/ perception frame]{
        \includegraphics[width=0.23\linewidth,height=0.14\linewidth]{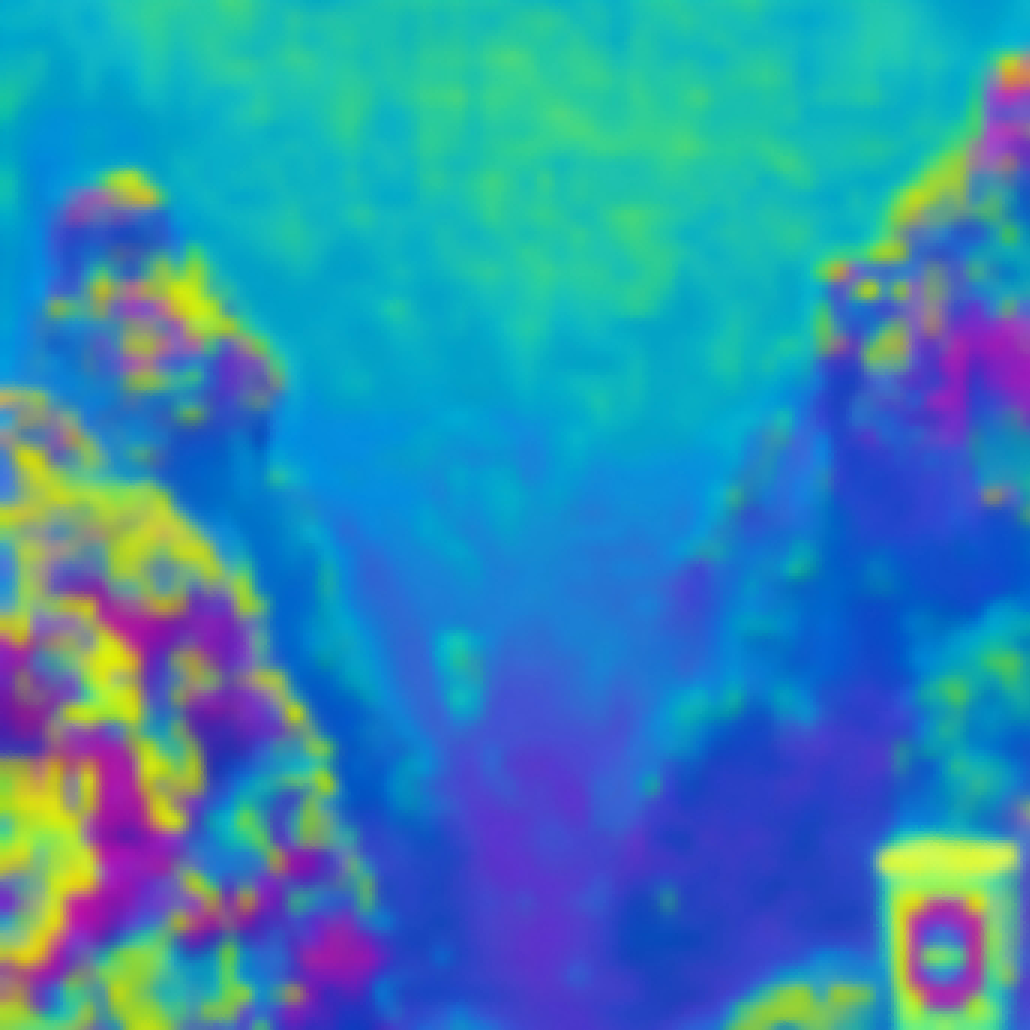}
    }
    \caption{Visualization of attention map and perception frame effects.}
    \label{fig:attention}
\end{figure*}
\noindent\textbf{Visualizing perception frames.} To better understand the role of perception frames, we visualize the attention responses of the stereo encoder with and without the proposed perception frames in Figure~\ref{fig:attention}. Without perception frames, the attention is scattered over degraded regions, especially around low-texture areas, strong color casts, and backscatter-corrupted boundaries, leading to less stable correspondence cues. In contrast, after introducing perception frames, the model produces more concentrated and structure-aware attention on object contours and reliable matching regions. This suggests that the learnable perception frames enrich the pseudo temporal context, absorb underwater priors during training, and guide cross-frame attention toward view-consistent structures.
\section{Conclusion}
We present \textbf{AquaStereo}, a perception-enhanced framework that improves underwater stereo matching by jointly addressing data scarcity and feature degradation. On the data side, we introduce a depth-conditioned diffusion pipeline that synthesizes geometry-faithful underwater stereo pairs, guided by a physics-inspired prompt pool encoding water types and image-formation cues to control global appearance. A lightweight left--right consistency module further enforces binocular structure and epipolar alignment during generation, narrowing the terrestrial--underwater gap. On the training side, cross-domain self-distillation transfers clean geometric cues from a frozen teacher to a student trained on underwater pairs with perturbations, while clean-branch supervision with shared pseudo targets prevents scale drift. On the model side, we fuse learnable perception frames encoded by a video backbone with high-level semantics from a strong image encoder to produce robust matching descriptors in turbid, low-texture conditions. Extensive zero-shot evaluations on UWStereo, FLSea, Squid, and TartanAir, together with ablations, demonstrate consistent gains in accuracy and stability. Overall, AquaStereo offers a practical recipe for reliable underwater stereo and motivates future work on stronger physics-guided generation, online adaptation, and real-time efficiency.

\section*{Acknowledgements}
This work was supported by the National Natural Science Foundation of China (62331006, 625B2026), and the Fundamental Research Funds for the Central Universities.

%
%
\bibliographystyle{splncs04}
\bibliography{main}
\end{document}